\documentclass{article}

\PassOptionsToPackage{numbers, compress}{natbib}

\usepackage[dblblindworkshop, final]{neurips_2026}

\usepackage[utf8]{inputenc}
\usepackage[T1]{fontenc}
\usepackage{hyperref}
\usepackage{url}
\usepackage{booktabs}
\usepackage{amsfonts}
\usepackage{nicefrac}
\usepackage{microtype}
\usepackage{xcolor}

\usepackage{amsmath}
\usepackage{graphicx}
\usepackage{subcaption}
\usepackage{multirow}

\graphicspath{{./img/}{./figures_corrected/}{./figures/}}

\title{What Words Keep of a Place: Zero-Shot Language\\
Reasoning for Cross-View Geo-Localization}

\workshoptitle{Physical World AI: Geometry, Characteristics, and Multimodal Sensing}

\author{%
  Ayesh Abu Lehyeh\textsuperscript{1},
  Jay Hwasung Jung\textsuperscript{1},
  Safwan Wshah\textsuperscript{1} \\\\
  \textsuperscript{1}Department of Computer Science, University of Vermont \\
  \texttt{Ayesh.Abulehyeh@uvm.edu} \quad
  \texttt{Jay-Hwasung.Jung@uvm.edu} \quad
  \texttt{Safwan.Wshah@uvm.edu}
}

\begin{document}

\maketitle

\begin{abstract}
Cross-view geo-localization is commonly solved as an image retrieval problem, matching a ground-level image against a database of satellite tiles through a jointly trained embedding. Such models are accurate, but they need large paired supervision and cannot show what evidence supports a match. In this paper, we study a different question: how much of this task can be solved through language alone? We prompt a multimodal large language model (MLLM) to describe each ground panorama and each satellite tile as structured text, and localize by comparing these descriptions. No component is trained. We evaluate on 9,826 VIGOR pairs from four U.S. cities, in three settings. First, the descriptions are faithful but not discriminative. They agree closely across the two views, yet ranking the full pool by description similarity almost never returns the correct tile ($0.39\%$ Recall@1). Second, we narrow the pool to ten neighboring tiles, as a coarse prior would do. The same descriptions now become useful: an MLLM judge that scores structural consistency doubles random ranking and matches a strong lexical baseline. It also states which fields of the two descriptions agree and which conflict, which an embedding distance cannot do, and which we see as a step toward interpretable localization. Third, we place the judge on a trained visual retriever. On the queries it ranks wrongly, reranking from images works, while reranking from our descriptions does not ($23.5\%$ against $10.7\%$ Recall@1). Scene structure survives the conversion into language, while the fine appearance detail needed to separate nearby places does not. Code and prompts are publicly available at \url{https://github.com/AyeshAbuLehyeh/GeoLingual}.
\end{abstract}
\section{Introduction}
\label{sec:introduction}

Cross-view geo-localization aims to find which satellite tile, out of a large georeferenced database, covers the location of a ground-level query image. The task matters whenever GPS is degraded~\cite{ubigtloc, spatially}, for example in dense urban canyons, or for an embodied agent that tracks its position from what it sees~\cite{geoflow}. Most current systems treat the problem as metric learning over image pairs~\cite{Vigor, geodtr, zhang2024geodtr+, deuser2023sample4geo}, training an encoder so that a panorama and its satellite tile lie close together in a shared embedding space. This works well on benchmarks, but it has two drawbacks. First, accuracy depends on large amounts of paired supervision and tuned negative sampling. Second, the decision is opaque. An embedding distance does not show which parts of the scene produced the match, a limitation in the safety-critical settings that motivate the task.

Recent work brings language into this pipeline, either by training text-to-satellite retrieval on MLLM-generated descriptions~\cite{ye2025cvgtext} or by using descriptions to rerank and explain visual candidates~\cite{dagda2025geovlm}. Our goal is different. Instead of training a language component and reporting the gain, we keep the pipeline zero-shot and study the description step itself. What does a structured verbal account of a scene keep, and what does it lose? We prompt a multimodal large language model (MLLM) to convert both views into machine-readable XML descriptions of landmarks, road topology and spatial relations. The same model, queried text-only, then reasons over these descriptions to choose a tile. Nothing is fine-tuned: the descriptions come from prompting, the baselines get no contrastive training, and the judge has never seen a labeled cross-view pair. We use the word reasoning in a narrow sense. The judge does not compute a similarity score. It compares named fields across the two descriptions, states which agree and which conflict, and commits to a verdict that a reader can check against the text it was given. The decision is auditable, which an embedding distance is not. We therefore see this work as a step toward interpretable cross-view geo-localization rather than as a new retrieval method, and we do not claim that the rationale faithfully reflects the model's internal computation (Section~\ref{subsec:limitations}). The underlying question is about representation. A panorama and an overhead tile record the same place through very different sensors, and a description is a third record that depends on neither. Asking whether it supports localization is asking which properties of a scene survive being put into words. Our field ablation gives a partial answer: road topology and geometry carry the signal, orientation cues do not.

The study proceeds in three steps. First, we confirm that the descriptions are consistent and semantically aligned across the two views (Section~\ref{subsec:res_lexical}). Second, we show that this alignment is not enough for retrieval: ranking every satellite description against a ground query gives a Recall@1 below $0.4\%$ for all three text representations. The failure is one of distinctiveness, not accuracy, so we restrict the problem to \textit{verification} over a small pool of geographically adjacent candidates, the situation a coarse prior would produce. Language reasoning becomes useful here, and states a rationale for each decision. Third, we place the MLLM on top of a pretrained retriever and rerank the queries it gets wrong. The outcome inverts. Reranking from images works, reranking from our descriptions does not improve on chance, and giving the model both modalities does not recover the difference. The limits here therefore come less from the reasoning step than from what the description step preserves.

In this work, our contributions can be summarized as threefold:
\begin{itemize}
    \item We propose structured spatial verification, which recasts candidate ranking as a pointwise LLM-as-a-judge task over pools of geographically adjacent hard negatives (Section~\ref{subsec:res_retrieval}, Table~\ref{tab:text}b). It reaches $21.0\%$ Recall@1 against a $10\%$ random baseline. This is ahead of dense embeddings and level with TF-IDF within sampling noise. It is also the only method here that explains itself: each decision names the fields that agree between the two views and the fields that conflict (Appendix~\ref{sec:appendix_qualitative}). A field ablation attributes most of the signal to road topology and geometry, while orientation cues alone fall to chance.
    \item We present a retrieve-then-rerank study on the queries a trained retriever gets wrong. Image-based listwise reranking recovers $23.52\%$ of these failures, while text-based ($10.66\%$) and fused image-plus-text reranking ($11.00\%$) stay close to chance (Section~\ref{subsec:res_rerank}, Figure~\ref{fig:rerank_bar}). This indicates that concatenating the two modalities in a prompt dilutes the stronger signal instead of complementing it.
    \item We give an account of when language helps in this task and when it does not, separating candidate sets that are geographically hard from those that are visually hard (Section~\ref{sec:discussion}). Two directions follow: training a verifier on our hard negative pools, and training the description model with a discriminability-aware objective.
\end{itemize}
\section{Related Works}
\label{sec:related_work}

\textbf{Cross-View Geo-Localization (CVGL).} CVGL matches a ground-level image against a georeferenced satellite database, and almost all methods solve it through learned visual similarity. Early work such as CVUSA~\cite{CVUSA}, CVM-Net~\cite{cvmnet} and SAFA~\cite{SAFA} compared ground and satellite features with CNN encoders and view-aware pooling. Later methods, including L2LTR~\cite{l2ltr}, DSM~\cite{DSM} and GeoDTR~\cite{geodtr, zhang2024geodtr+}, improved accuracy with transformer-based matching and with geometric or domain alignment. The VIGOR dataset~\cite{Vigor} made the setting more realistic. It dropped the assumption that a panorama sits at the center of its tile, and allowed one tile to cover several panoramas. This is the version of the problem we work in. In recent systems the largest gains come from the training objective rather than the backbone, combining contrastive losses with hard-negative sampling on modern convolutional encoders~\cite{deuser2023sample4geo, convnext, auxgeo}. We use a model of this family as the retrieval front-end in Section~\ref{subsec:vision_rerank}. All of them depend on paired supervision and tuned negative mining, and none can be inspected after a match is made.

\textbf{Language-based cross-view localization.} A recent line of work localizes from natural-language scene descriptions rather than from image features alone, and it is the closest to ours. CVG-Text pairs street views with MLLM-generated descriptions across three cities and trains a text-to-reference retrieval model, CrossText2Loc, reporting gains in accuracy and explainability~\cite{ye2025cvgtext}. GeoVLM uses vision-language descriptions to rerank cross-view candidates and justify the choice~\cite{dagda2025geovlm}, and GeoX-Bench evaluates large multimodal models on cross-view correspondence at scale~\cite{zheng2026geoxbench}. These papers ask how well a language-based pipeline can perform, and answer by training one. We ask a complementary question. With nothing trained, what does the description step keep and what does it discard? Our reranking experiment is informative here, since it identifies a regime in which description-based reranking carries no signal at all: the queries a trained retriever ranks incorrectly.

\textbf{Multimodal large language models.} Models such as CLIP~\cite{DBLP:journals/corr/abs-2103-00020}, BLIP-2~\cite{li2023blip2bootstrappinglanguageimagepretraining}, LLaVA~\cite{liu2023llava} and Qwen3~\cite{qwen} make this approach possible. They name relations such as "a road curving left behind a row of trees" that conventional backbones encode only implicitly, and they do so in readable text.

\textbf{LLM-as-a-judge.} A separate line of work uses language models to \textit{evaluate} or \textit{rank} content in place of human judgment or embedding similarity~\cite{zheng2023judging}. In information retrieval, rerankers such as RankGPT~\cite{sun2023chatgpt} prompt a model to order candidates against a query, and outperform embedding similarity when compositional judgments matter more than topical overlap. Related work asks a model to audit its own generations for unsupported content~\cite{manakul2023selfcheckgpt}, and we adapt this as a grounding check in Section~\ref{subsec:verification}. We apply the judging paradigm to CVGL. Instead of ranking candidate satellite descriptions by cosine similarity, we prompt the same MLLM, text-only, to assess structural and spatial consistency between a ground query and each candidate. The usual application of LLM-as-a-judge targets open-ended response quality, and it is known to suffer from position, verbosity and self-enhancement biases. Our task is narrower, since the properties at issue are checkable in principle: road topology, orientation and landmark correspondence.
\section{Methodology}
\label{sec:methodology}

\begin{figure}[t]
    \centering
    \includegraphics[width=0.92\linewidth]{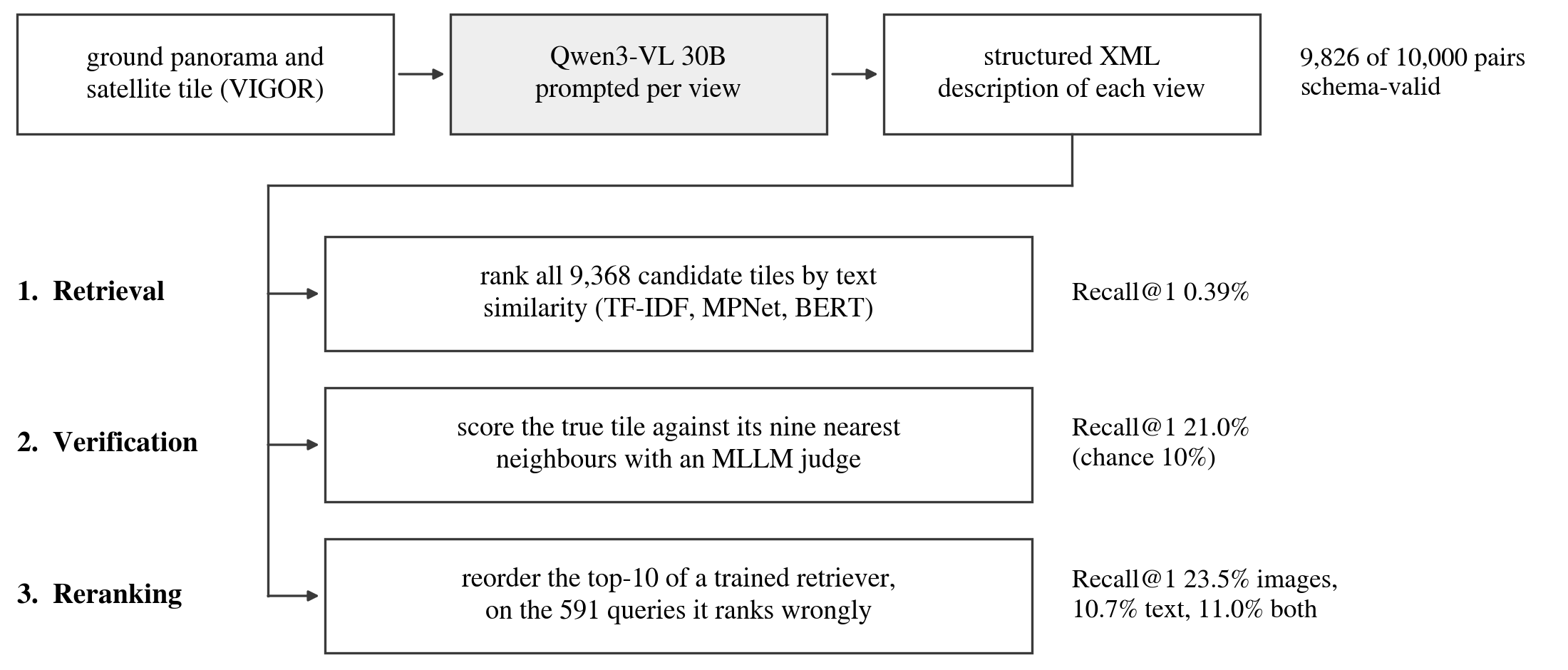}
    \caption{Overview of the study. A prompted MLLM converts both views into structured XML descriptions. The same descriptions are then used in three settings: retrieval over the full pool, pointwise verification inside a small geographic neighborhood, and listwise reranking of the top-10 of a trained visual retriever. No component is fine-tuned. Setting 3 is scored on the 591 queries whose visual top-1 was wrong.}
    \label{fig:pipeline}
\end{figure}

\subsection{Dataset}
\label{subsec:dataset}

We use \textbf{VIGOR}~\cite{Vigor}, which pairs 360$^{\circ}$ ground-level panoramas from Google Street View with satellite tiles covering the same area, together with GPS metadata. It contains roughly 90{,}000 panoramas and 10{,}000 satellite regions across four U.S. cities, and we work with 2{,}500 paired samples per city to limit inference cost (Appendix~\ref{sec:appendix_figures}).

This subsampling affects how our numbers should be read. We take the first 2{,}500 rows per city of the official same-area test split, and neighboring panoramas can share a tile, so the reference pool holds 2{,}361 to 2{,}401 unique tiles per city, or 9{,}368 in total. That is roughly ten times smaller than the full VIGOR same-area pool. Absolute Recall@$K$ values here are therefore \textit{not} comparable to figures reported on the full benchmark. Every method we evaluate uses the identical reduced pool, so comparisons \textit{within} the paper remain valid, and Section~\ref{subsec:res_rerank} quantifies the effect.

\subsection{Textual Description Generation and Prompt Design}
\label{subsec:prompt_design}

No component is trained, so the prompts carry the work that training would otherwise do. Generic captioning was not sufficient in preliminary trials, since it produced text that could not separate one street corner from the next. We therefore wrote one prompt per modality. The satellite prompt enforces an allocentric, map-based perspective, and extracts cardinal orientations, roof fingerprints and linear sequences of static objects. The ground prompt enforces an egocentric perspective, and focuses on road topology, landmark visibility front and rear, and distinctive anchors. Both prompts constrain the output to a strict XML schema, which keeps the descriptions machine-readable by grouping cues into fixed categories. The schema was hand-designed and kept fixed for all experiments, so our findings describe this particular representation rather than language in general. Appendix~\ref{sec:appendix_prompts} gives the full prompts and schemas.

We use Qwen 3 VL 30B A3B Instruct~\cite{qwen} on NVIDIA H200 GPUs. The same model performs description generation, verification and reranking, always by prompting alone with greedy decoding; Section~\ref{subsec:limitations} discusses the resulting judge-generator overlap.

\subsection{Natural Language Processing and Analysis Framework}
\label{subsec:nlp_analysis}

\textbf{XML parsing and structural repair.} Generative models occasionally emit malformed markup. We therefore implemented a rule-based parser that validates every description against its schema, normalizing tag casing, closing unclosed tags, discarding overlapping blocks and stripping commentary outside the root element. Keeping only the samples that pass validation after repair recovers 9{,}826 valid pairs from the 10{,}000 generated samples, or $98.3\%$, balanced across cities. The remaining $1.7\%$ are excluded (Appendix~\ref{sec:appendix_impl}).

\textbf{Lexical distribution.} We ran a Bag-of-Words analysis on the parsed content to check whether the two views describe the same physical entities. We lemmatize with \texttt{spaCy}, keep only nouns and proper nouns, and preserve domain bigrams such as "parking lot" as single tokens. We then compare the frequency of semantic anchors (e.g., \textit{tree}, \textit{intersection}, \textit{lane}) across modalities.

\textbf{Semantic embedding and zero-shot retrieval.} We encoded the cleaned descriptions with a pre-trained BERT~\cite{BERT} model, mean-pooling the final layer, and scored each pair $(G_i, S_i)$ by cosine similarity, $\text{Sim}(G_i, S_i) = \mathbf{v}_{G_i} \cdot \mathbf{v}_{S_i} / (\|\mathbf{v}_{G_i}\| \|\mathbf{v}_{S_i}\|)$. For every ground description we rank all candidate satellite descriptions in the reduced pool and report Recall@$K$. Raw BERT embeddings are poorly suited to sentence similarity, so we treat them as a reference point and add two further representations on the identical pool: a sentence-transformer trained for semantic similarity (\texttt{all-mpnet-base-v2})~\cite{reimers2019sentence, song2020mpnet}, and a sparse TF-IDF vectorizer over the same lemmatized tokens. None of the three receives contrastive fine-tuning, and none is a modern retrieval-tuned encoder. We return to this scope limit in Section~\ref{subsec:limitations}.

\subsection{Structured Spatial Verification as an LLM-as-a-Judge Task}
\label{subsec:verification}

The retrieval formulation asks a representation to separate the true tile from thousands of candidates at once. Deployed systems rarely face that problem in its raw form, since a coarse prior from GPS, dead reckoning or a first-stage retriever narrows the search to a small neighborhood. We therefore reframe candidate ranking as a pointwise LLM-as-a-judge task inside such a neighborhood, as a verification step that follows a geographic prior rather than a replacement for retrieval. For a ground query and each candidate satellite description, the same MLLM is queried text-only and without images. It scores spatial consistency on a 0-10 scale and issues a MATCH or NO\_MATCH verdict with a short rationale. The judge is conditioned on the structured fields already extracted for both views: road topology, spatial layout, orientation cues and distinctive anchors for the ground view, and cardinal orientation, road geometry, linear sequence and distinctive anchors for the satellite view. Candidates are ranked by this score.


The candidate pools are deliberately difficult. Each contains the true tile and the nine geographically nearest wrong tiles in the same city, the nearest lying 55 to 61 meters away on average. These are the cases a coarse prior cannot resolve. We evaluate on 600 queries, and the three baselines of Section~\ref{subsec:nlp_analysis} score the identical pools, so any difference reflects the ranking mechanism rather than the candidate set. The baselines read the full parsed description, while the judge sees only the fields listed above, so the judge is not favored by having more input. Random ranking gives $10\%$ Recall@1 and $50\%$ Recall@5, our reference point throughout. Two auxiliary experiments probe this setting. A \textit{field ablation} repeats the task on 120 queries per condition while exposing only part of the schema. A \textit{grounding audit} re-presents 356 \texttt{distinctive\_anchors} claims from 80 panoramas with their source images, labeling each GROUNDED, PARTIAL or HALLUCINATED. Appendix~\ref{sec:appendix_ablation} details both.

\subsection{Vision-Retrieval-Augmented Reranking}
\label{subsec:vision_rerank}

The verification setting assumes some external prior has produced the candidate pool. In our final experiment that prior is a real visual system, and we ask whether language reasoning adds anything on top of it. We use a pretrained cross-view retrieval checkpoint from AuxGeo~\cite{auxgeo}, a ConvNeXt-base encoder trained contrastively with hard-negative mining on the VIGOR same-area split in the manner of Sample4Geo~\cite{deuser2023sample4geo}, and run it over our reduced pool to obtain, for all 9{,}826 queries, the top-10 candidate tiles and the rank of the true match. Because the retriever is close to ceiling here, reranking can only matter where it fails. We define the \textit{hard subset} as the queries whose visual top-1 is wrong, and report \textit{recovery rate}, the fraction of those queries whose true tile the reranker places first.

The reranker is listwise rather than pointwise. All ten candidates appear in a single labeled prompt, and the model outputs a full ranking instead of an independent score per candidate. This follows standard practice in the LLM reranking literature~\cite{sun2023chatgpt}, where comparing candidates outperforms judging each in isolation, and it suits near-duplicates whose differences are visible only by comparison. Holding the queries, pools and prompt structure fixed, we compare three reranker inputs: \textbf{text only}, the structured descriptions of the query and all ten candidates; \textbf{image only}, the raw panorama and the ten raw tiles; and \textbf{image and text}, both concatenated in the same prompt. The variants run on identical query lists with the same seed, which makes their comparison a clean ablation of input modality.
\section{Results}
\label{sec:results}

\subsection{Lexical Distribution and Semantic Alignment}
\label{subsec:res_lexical}

The two modalities are consistent in what they mention. For every anchor we track, the frequency is close between the ground and satellite views within a city, so the prompts extract comparable content across the viewpoint gap. The absolute level is more revealing. The dominant anchors are near saturation, and "building" appears in more than $90\%$ of descriptions in both views (Appendix~\ref{sec:appendix_figures}). A term occurring in nearly every description carries little discriminative value, so consistency and low discriminability are two readings of one measurement.

The embedding analysis points in the same direction. Paired descriptions reach a mean BERT cosine similarity of $0.8748$, tightly concentrated around that mean (Figure~\ref{fig:cosine} in the appendix). High similarity on true pairs is useful only if wrong pairs score lower, and this is where it breaks down: off-diagonal entries frequently exceed the diagonal, and mismatched pairs fall in the same narrow band as correct ones.

\subsection{From Retrieval to Verification}
\label{subsec:res_retrieval}

Panel (a) of Table~\ref{tab:text} measures the lack of distinctiveness directly. The best of the three representations reaches $0.39\%$ Recall@1 over the reduced pool, and none exceeds $3.2\%$ Recall@20. Two factors contribute. The first is a vocabulary gap: the same entity is called a "commercial storefront" from the ground and a "flat grey rectangular building" from above. The second is semantic homogeneity, since frequent terms such as "road" and "building" dominate the similarity and bury whatever would identify a specific place. Sparse TF-IDF outperforms both neural representations, consistent with the reported anisotropy of un-finetuned BERT embeddings~\cite{li2020sentence, ethayarajh2019contextual}. When almost all shared vocabulary is generic, an exact match on a rare token is the most informative signal available. These numbers describe our descriptions under zero-shot matching. They are not an upper bound on text-based localization, where trained models do considerably better~\cite{ye2025cvgtext}.

Panel (b) turns to the verification formulation of Section~\ref{subsec:verification}, where all four methods rank the same ten-candidate pools. Structured verification reaches $21.0\%$ Recall@1 with an MRR of $0.419$, nearly double raw BERT and ahead of the sentence-transformer on every metric. It is not ahead of TF-IDF, whose $20.0\%$ lies inside the $\pm 3.3$ point interval. The tie is as informative as the improvement. Explicit reasoning over structured fields improves on dense embedding similarity, but not on a well-built lexical baseline, because exact term overlap remains effective even against geographically close negatives. What the judge adds at comparable accuracy is an explanation. Every decision comes with a short rationale naming which structured fields agree across the two views and which conflict, for example a shared T-intersection against a roof layout that does not match. The baselines cannot do this: a retriever returns a distance and TF-IDF a set of overlapping tokens, so neither shows what the match rests on. An operator can check the rationale against the two descriptions and reject a match when the evidence is weak. Appendix~\ref{sec:appendix_qualitative} reproduces three decisions in full, and in the false accept the rationale itself exposes the error, since it claims an agreement that appears in only one description. Matching the strongest baseline while explaining itself is the claim we defend, and a better outcome than panel (a) suggests is possible from text.

\begin{table}[t]
    \centering
    \small
    \caption{Text-based localization in two formulations. Panel (a) ranks all satellite descriptions in the reduced pool. Panel (b) ranks ten-candidate pools of the true tile and its nine nearest wrong tiles, identical across methods. Values are percentages except MRR. The $95\%$ binomial interval on Recall@1 is $\pm 0.12$ points in panel (a) and $\pm 3.3$ points in panel (b), so the top two rows of panel (b) are not separated by these data and are left unbolded.}
    \label{tab:text}
    \begin{tabular}{lcccc}
        \toprule
        \multicolumn{5}{l}{\textit{(a) Zero-shot text-to-text retrieval, full reduced pool ($n = 9{,}826$)}} \\
        & \textbf{R@1} & \textbf{R@5} & \textbf{R@10} & \textbf{R@20} \\
        \midrule
        Raw BERT (\texttt{bert-base-uncased}, mean-pooled) & 0.02 & 0.16 & 0.26 & 0.49 \\
        Sentence embedding (\texttt{all-mpnet-base-v2})    & 0.14 & 0.53 & 0.77 & 1.15 \\
        TF-IDF (lemmatized nouns, preserved bigrams)       & \textbf{0.39} & \textbf{1.16} & \textbf{1.94} & \textbf{3.17} \\
        \midrule
        \multicolumn{5}{l}{\textit{(b) Ranking hard negative pools, matched candidates ($n = 600$)}} \\
        & \textbf{R@1} & \textbf{R@3} & \textbf{R@5} & \textbf{MRR} \\
        \midrule
        MLLM structured verification                    & 21.0 & 50.3 & 68.2 & 0.419 \\
        TF-IDF                                          & 20.0 & 50.2 & 70.0 & 0.412 \\
        Sentence embedding (\texttt{all-mpnet-base-v2}) & 14.5 & 38.8 & 61.7 & 0.350 \\
        Raw BERT                                        & 11.7 & 34.2 & 54.8 & 0.313 \\
        Random ranking (expected)                       & 10.0 & 30.0 & 50.0 & 0.293 \\
        \bottomrule
    \end{tabular}
\end{table}

The field ablation (Appendix~\ref{sec:appendix_ablation}) shows where this signal comes from. Exposing only road topology and geometry matches the full schema to within sampling noise, at $17.5\%$ against $20.8\%$ Recall@1 at $n = 120$ with an interval of $\pm 7$ points. Orientation cues alone fall to $5.8\%$, at or below the random baseline. Road geometry survives the change of viewpoint, while orientation cues are a coarse compass reading shared by most tiles in a city grid. Notably, the landmark fields carry no more signal than topology. Street names appear in those fields when the model recognizes a place, so this limits how much of the result could rest on memorized geography (Appendix~\ref{sec:appendix_qualitative}). A grounding self-audit over 356 anchor claims labels $92.1\%$ verifiable and $4.8\%$ hallucinated. The auditing model also wrote the claims, so the first figure is optimistic and the second a lower bound (Section~\ref{subsec:limitations}).

\subsection{Vision Retrieval and MLLM Reranking}
\label{subsec:res_rerank}

The pretrained retriever reaches $93.99\%$ Recall@1 and $99.17\%$ Recall@10 on our pool ($n = 9{,}826$). Its reported Recall@1 on the full VIGOR same-area test set is $80.43\%$, so the higher figure follows from a pool ten times smaller. Only 591 queries, about $6\%$, have an incorrect visual top-1 to repair, and we evaluate all three variants on this subset. The true tile is in the top-10 for about $86\%$ of them, so a reranker can reach at most $86\%$ and the effective chance level is $8.6\%$ rather than $10\%$.

\begin{figure}[t]
    \centering
    \includegraphics[width=0.42\linewidth]{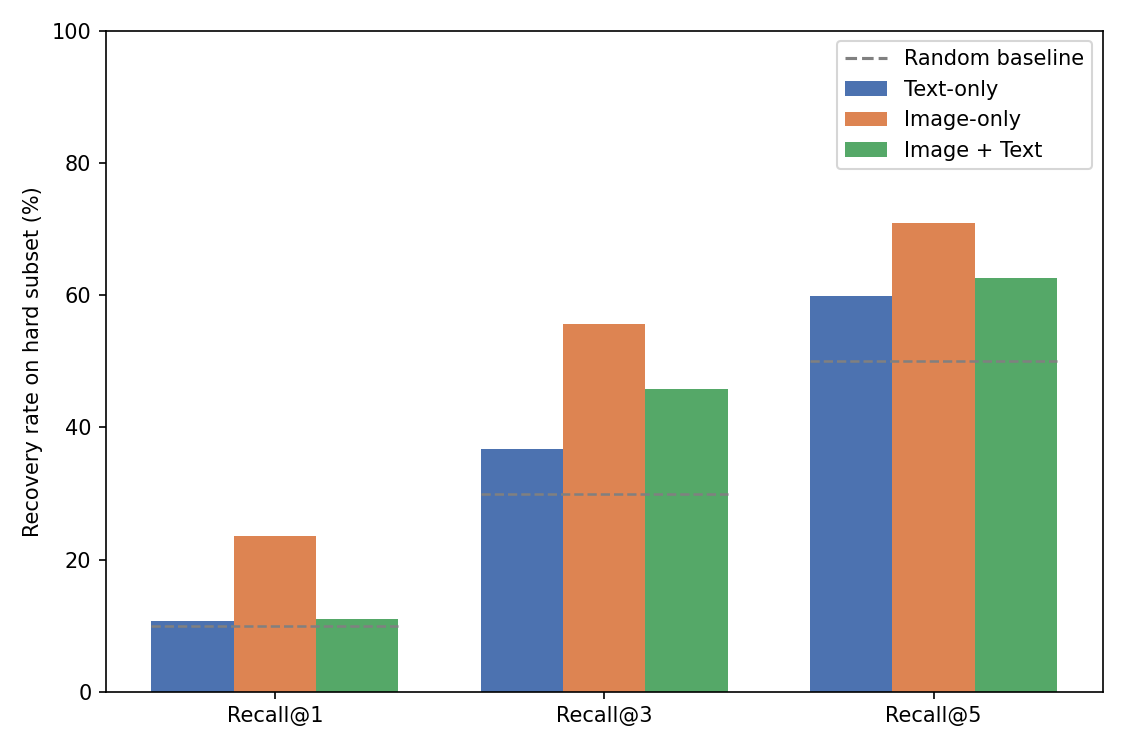}
    \caption{Listwise reranking of the retriever's top-10 on the subset where its own top-1 was incorrect ($n = 591$, identical queries and pools across variants). The true tile is absent from the pool for about $14\%$ of these queries, so the effective chance level is $8.6\%$ at Recall@1, and only the image-only variant separates clearly from it.}
    \label{fig:rerank_bar}
\end{figure}

Figure~\ref{fig:rerank_bar} reports the outcome, and the pattern is asymmetric. Text-only reranking recovers $10.66\%$ and image-plus-text $11.00\%$, both within about two points of chance. Image-only reaches $23.52\%$ ($\pm 3.4$ points), more than twice the text-only figure. The combined variant does beat text alone further down the ranking, at $45.85\%$ against $36.72\%$ for Recall@3, so adding images helps recall without restoring top-1 precision. The rank histograms in Appendix~\ref{sec:appendix_figures} agree: image-only is sharply peaked at rank 1, while the other two are close to flat.

These are conditional rates. If reranking never disturbed an already-correct top-1, image-only reranking would raise end-to-end Recall@1 from $93.99\%$ to $95.40\%$, text-only to $94.63\%$ and the combined variant to $94.65\%$, with a ceiling of $99.16\%$. We report these as upper bounds, since we have not measured how often reranking damages a correct top-1.

Two observations follow. First, text-only reranking is close to chance here, even though the same descriptions supported clearly above-chance verification in Table~\ref{tab:text}. What differs is the kind of difficulty: the candidates are geographically close in one case and visually confusable in the other. The tiles a contrastively trained model cannot separate are near-duplicates, and their descriptions collapse onto the generic vocabulary of Section~\ref{subsec:res_lexical}. Second, presenting both inputs together does not simply fail to help: it returns Recall@1 to the text-only level while keeping part of the image advantage lower in the ranking. The likely reason is that near-identical generic text competes for attention with the sharper visual evidence. In the cases we inspected, the true tile and the selected distractor were described with the same intersection type and roof colors. We tested one fusion strategy, so this argues against naive concatenation and not against multimodal fusion in general. Together with Table~\ref{tab:text}, these results show where language reasoning helps: when candidates are geometrically distinct but lexically confusable, and not when they are visually near-identical.
\section{Discussion}
\label{sec:discussion}

The three experiments point to a limit of the representation rather than of the reasoning applied to it. A description records what language names easily, such as road topology and coarse spatial relations, and omits what it names poorly, such as roof color, texture and precise footprints. Verification over geographically adjacent candidates succeeds because those candidates usually differ in what the text keeps. Reranking the retriever's mistakes fails because they differ mainly in what it omits. There is also a practical asymmetry between the two representations. A retriever returns a distance, and nothing about that number can be argued with. Our judge returns an argument that can be checked against the two descriptions, so a wrong match can be traced to the field that misled it. This holds even when the two are equally accurate. Language is therefore better used as a channel for structure and explanation than as the measurement the decision rests on. All of this holds in the zero-shot setting we chose. We measure the content of prompted descriptions, not an upper bound on what a trained system could extract.

\subsection{Limitations}
\label{subsec:limitations}

\begin{itemize}
    \item \textbf{Generation and reasoning are not fully separated.} Our evidence that the description step is the binding constraint is indirect. Descriptions that support verification fail on visually confusable candidates, while images on those candidates do not. The prompt format also changes between the two settings, from pointwise to listwise, so we cannot attribute the whole gap to description content. An oracle-description test, giving the judge descriptions known to contain a distinguishing cue, would separate the two effects and is the next experiment we plan.
    \item \textbf{Scope of the representation.} The schema is hand-designed, so our claims concern this representation and this model rather than language in general. Panel (a) of Table~\ref{tab:text} likewise bounds zero-shot matching over our descriptions, not text-based localization as such, and retrieval-tuned encoders would likely do better.
    \item \textbf{Self-audit and world knowledge.} Both the verifier and the grounding audit reuse the model that wrote the descriptions. For verification this is milder, since the judge compares two descriptions produced in independent calls. In the audit the model assesses its own output, the self-preference case described in the judging literature~\cite{zheng2023judging}, so $4.8\%$ hallucinated claims should be read as a lower bound. Appendix~\ref{sec:appendix_qualitative} shows two forms of the problem: a rationale citing an agreement present in neither description, and a verdict resting on recognized street names. The rationales are inspectable, but not guaranteed faithful.
    \item \textbf{Partial overlap between the views.} VIGOR panoramas are not centered on their tiles, so two faithful descriptions of one location can disagree. Our judge compares them as if they described the same content, which produces avoidable false rejections (Appendix~\ref{sec:appendix_qualitative}).
    \item \textbf{Evaluation scope.} The reduced pool inflates absolute recall for every method, and single runs at $n = 600$ and $n = 120$ leave intervals of roughly $\pm 3$ and $\pm 7$ points, which is why we treat close orderings as ties. Reranking is reported on the hard subset, with end-to-end figures given only as upper bounds.
\end{itemize}

\subsection{Future Work}
\label{subsec:future}

Our results point to two separate bottlenecks, and both are worth naming.

\textbf{Contrastive alignment and a trained verifier.} The retriever we use owes its accuracy to contrastive training with hard-negative mining, so the gap to a zero-shot judge is plausibly a training-signal gap rather than a capability gap. A natural next step is a projection trained to pull matching description embeddings together and push geographically near wrong tiles apart, using the pools from Section~\ref{subsec:verification} as labeled data.

\textbf{Discriminability-aware description generation.} Training the judge would not remove the second bottleneck. Text-only reranking landing close to chance suggests that the information needed to separate near-duplicate tiles is absent from the descriptions, not present but poorly used. No prompting strategy can recover what was never written down. Addressing this means training the description model with an objective that rewards text making the correct tile identifiable rather than merely accurate, which is closer to referring-expression generation than to caption fine-tuning. Given our fusion result, neither direction should concatenate modalities; an explicit fusion design and an image-only ablation would be needed.
\section{Conclusion}
\label{sec:conclusion}

In this paper, we studied what a structured linguistic description keeps and loses for cross-view geo-localization, using pretrained models zero-shot with prompts as the only degree of freedom. MLLM-generated descriptions are consistent and faithful across the viewpoint gap, but this does not make them distinctive, and ranking a large pool by description similarity almost never returns the correct tile. Once a coarse geographic prior narrows the pool, the same descriptions support verification about as accurate as a strong lexical baseline that, unlike any retrieval score, explains each decision in terms a person can check. This does not extend to candidates chosen for visual confusability. On the queries a trained retriever ranks wrongly, a reranker reading our descriptions performs close to chance, while one reading the images recovers a substantial share. We therefore see the verification stage as a step toward interpretable, reasoning-based cross-view geo-localization, and not as a replacement for visual retrieval. Two cautions follow. Concatenating a weaker language channel with a stronger visual one cost us most of the advantage of the stronger one, and a rationale that can be read is not always one that can be trusted.

\begin{ack}
This research was supported by the U.S. National Science Foundation under Grant No. 2119485.

The authors declare no competing interests.
\end{ack}

\bibliographystyle{unsrtnat}
\bibliography{references}

\clearpage
\appendix
\begin{center}
    {\LARGE\bfseries Appendix}
\end{center}
\vspace{0.5em}

The appendix is organized as follows. Appendix~\ref{sec:appendix_prompts} gives the two system prompts verbatim, which specify the description stage completely. Appendix~\ref{sec:appendix_figures} collects the supporting figures referenced from the main text. Appendix~\ref{sec:appendix_ablation} reports the field ablation and the grounding audit. Appendix~\ref{sec:appendix_qualitative} shows three verifier decisions in full, one correct and two wrong. Appendix~\ref{sec:appendix_impl} records implementation details.

\section{System Prompts for Description Generation}
\label{sec:appendix_prompts}

Below are the exact system prompts, reproduced verbatim, that generated every description used in the paper. Since prompting is the only degree of freedom in a zero-shot pipeline, these two blocks specify the description stage completely.

\subsection{Ground-Level Panorama Prompt}
Issued to the MLLM for $360^{\circ}$ equirectangular street-view images. It enforces an egocentric perspective and treats the image geometry as a continuous sphere.

\begin{verbatim}
You are a multimodal model specialized in 360° equirectangular street-view 
localization and navigation.

TASK:
Analyze the provided 360° panoramic image. You must interpret the image 
geometry as follows:
1. CENTER of image = Forward Heading (0°).
2. LEFT pixel coordinates = Left side, wrapping to Behind (-180°).
3. RIGHT pixel coordinates = Right side, wrapping to Behind (+180°).

OUTPUT FORMAT (XML):
<ground_scene>
  <road_topology_360>
    Describe the road network continuity. Explicitly link the road seen in 
    the Center (Front) with the road seen at the Edges (Back). 
    (e.g., "Straight road running North-South," "T-intersection ahead").
  </road_topology_360>
  <sector_forward>
    (Center View) Describe landmarks and geometry visible directly ahead 
    (the heading direction).
  </sector_forward>
  <sector_backward>
    (Edge Views) Describe the scene directly behind the camera position 
    (where the left and right edges of the image meet).
  </sector_backward>
  <spatial_layout>
    Describe the linear sequence of structures along the road boundaries 
    relative to the camera's path.
    - Left Side Sequence: (e.g., "Park -> Bus Stop -> White House")
    - Right Side Sequence: (e.g., "Commercial shops -> Gap -> Parking lot")
  </spatial_layout>
  <road_markings>
    Specific details of the road surface for precise positioning 
    (e.g., "double yellow center line," "white stop bar ahead").
  </road_markings>
  <orientation_cues>
    Visual evidence used to infer compass direction 
    (e.g., "Strong shadows casting to the right relative to heading").
  </orientation_cues>
  <distinctive_anchors>
     High-entropy, unique objects visible in ANY direction 
     (e.g., "Red fire hydrant on right curb," "Blue billboard").
  </distinctive_anchors>
  <environmental_context>
    Weather, season, and lighting conditions 
    (e.g., "Winter leafless trees," "Overcast sky").
  </environmental_context>
</ground_scene>

IMPORTANT:
- Analyze the image as a continuous sphere, not a flat 2D plane.
- Be precise with `road_markings` as they are critical.
- Only output XML format.
\end{verbatim}

\subsection{Satellite Image Prompt}
Issued to the MLLM for aerial satellite imagery. It enforces an allocentric perspective and directs the model to permanent map features rather than the ground camera location.

\begin{verbatim}
You are a multimodal model specialized in aerial geospatial analysis 
and cartography.

TASK:
Analyze ONLY the satellite image and generate a precise, structured 
semantic description focused on geolocalization cues.

OUTPUT FORMAT (XML):
<satellite_scene>
  <distinctive_anchors>
    Identify unique, high-entropy landmarks that act as "fingerprints" 
    (e.g., specific road intersections, unique building footprints).
  </distinctive_anchors>
  <roof_fingerprints>
    Describe the specific visual properties of rooftops which correlate 
    to building types (e.g., "flat grey industrial roof").
  </roof_fingerprints>
  <linear_sequence>
    Describe the sequence of objects along the road boundaries 
    (e.g., "continuous row of dense trees," "commercial storefronts").
  </linear_sequence>
  <cardinal_orientation>
    Define the road azimuth and alignment 
    (e.g., "road runs strictly North-South").
  </cardinal_orientation>
  <road_geometry>
    Detailed road topology (e.g., "2-lane road widening to 4 lanes," 
    "presence of median strip," "cul-de-sac").
  </road_geometry>
  <shadow_analysis>
    Describe shadow length and direction to infer building height 
    (e.g., "long shadows cast to the North indicating tall structures").
  </shadow_analysis>
</satellite_scene>

IMPORTANT:
- Analyze ONLY from the aerial/allocentric perspective.
- Do NOT infer the ground camera position.
- Focus on permanent features (roads, buildings) over transient ones.
- Only output XML format.
\end{verbatim}

\section{Additional Figures and Lexical Distributions}
\label{sec:appendix_figures}

\begin{figure}[h!]
    \centering
    \includegraphics[width=0.95\linewidth]{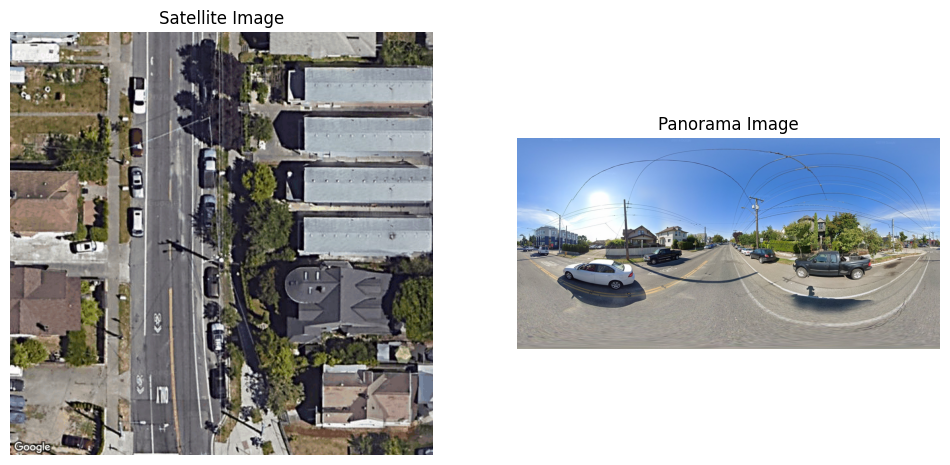}
    \caption{The two modalities in VIGOR. The panorama is not centered on its satellite tile, and one tile can cover several panoramas, which is the setting all our retrieval experiments use.}
    \label{fig:vigor_overview}
\end{figure}

\begin{figure}[h!]
    \centering
    \begin{subfigure}{0.46\linewidth}
        \centering
        \includegraphics[width=\linewidth]{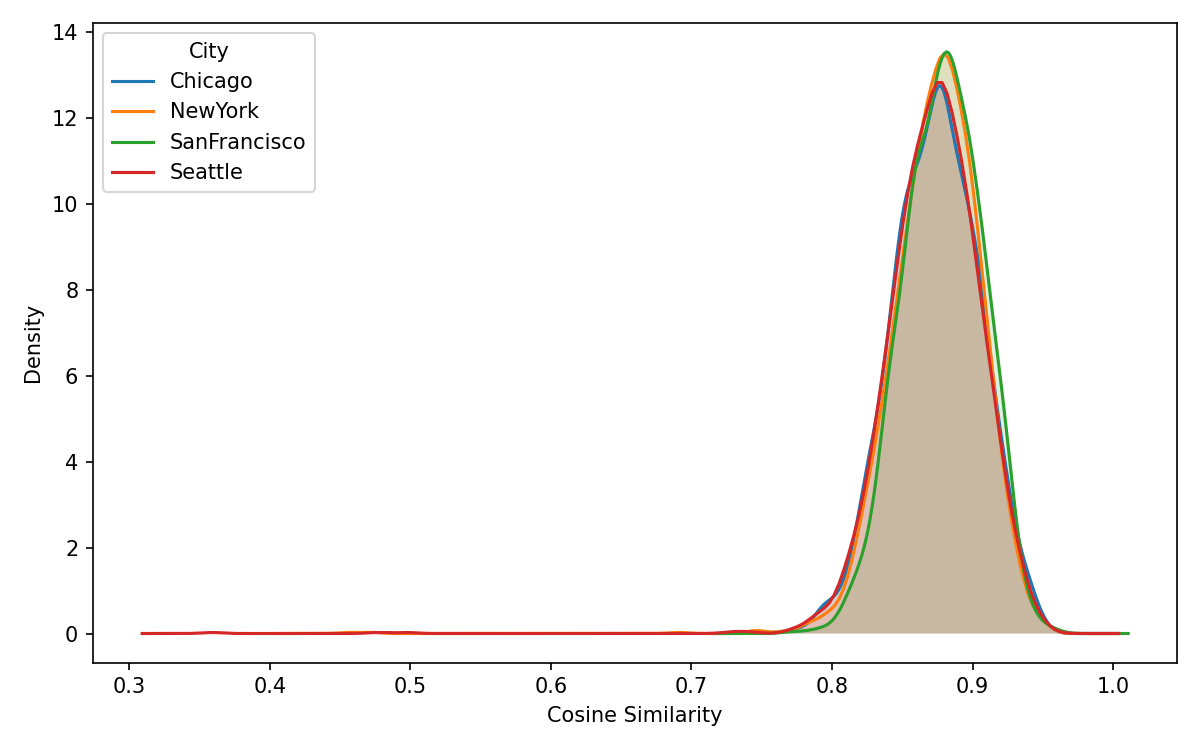}
        \caption{Similarity distribution (KDE).}
    \end{subfigure}
    \hfill
    \begin{subfigure}{0.46\linewidth}
        \centering
        \includegraphics[width=\linewidth]{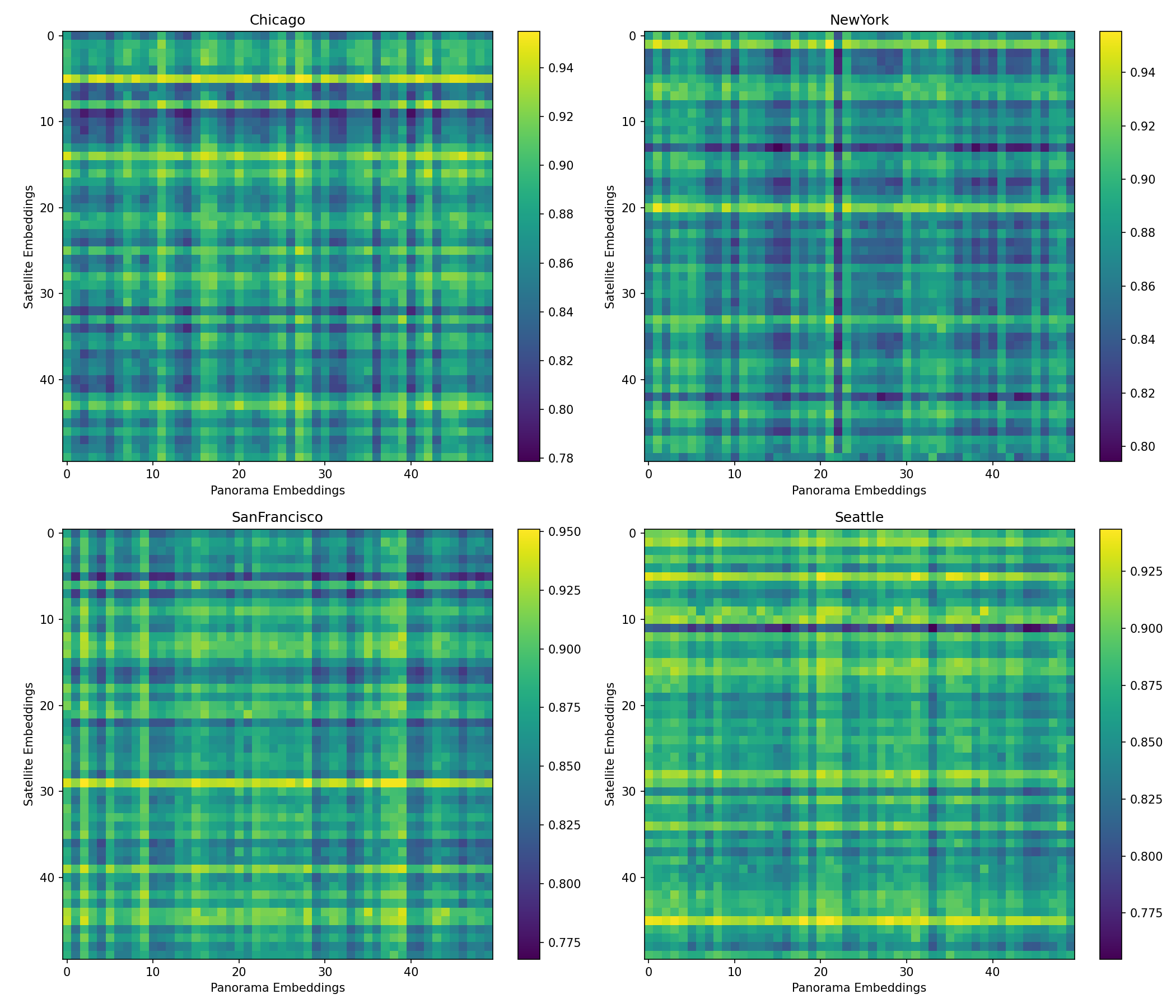}
        \caption{Pairwise similarity heatmap.}
    \end{subfigure}
    \caption{BERT cosine similarity between paired satellite and panorama descriptions. True pairs are highly similar (mean $0.8748$, standard deviation $0.0311$), but the distribution is narrow and several off-diagonal entries exceed the diagonal, so high similarity does not imply distinguishability.}
    \label{fig:cosine}
\end{figure}

\begin{figure}[h!]
    \centering
    \begin{subfigure}{0.49\linewidth}
        \centering
        \includegraphics[width=\linewidth]{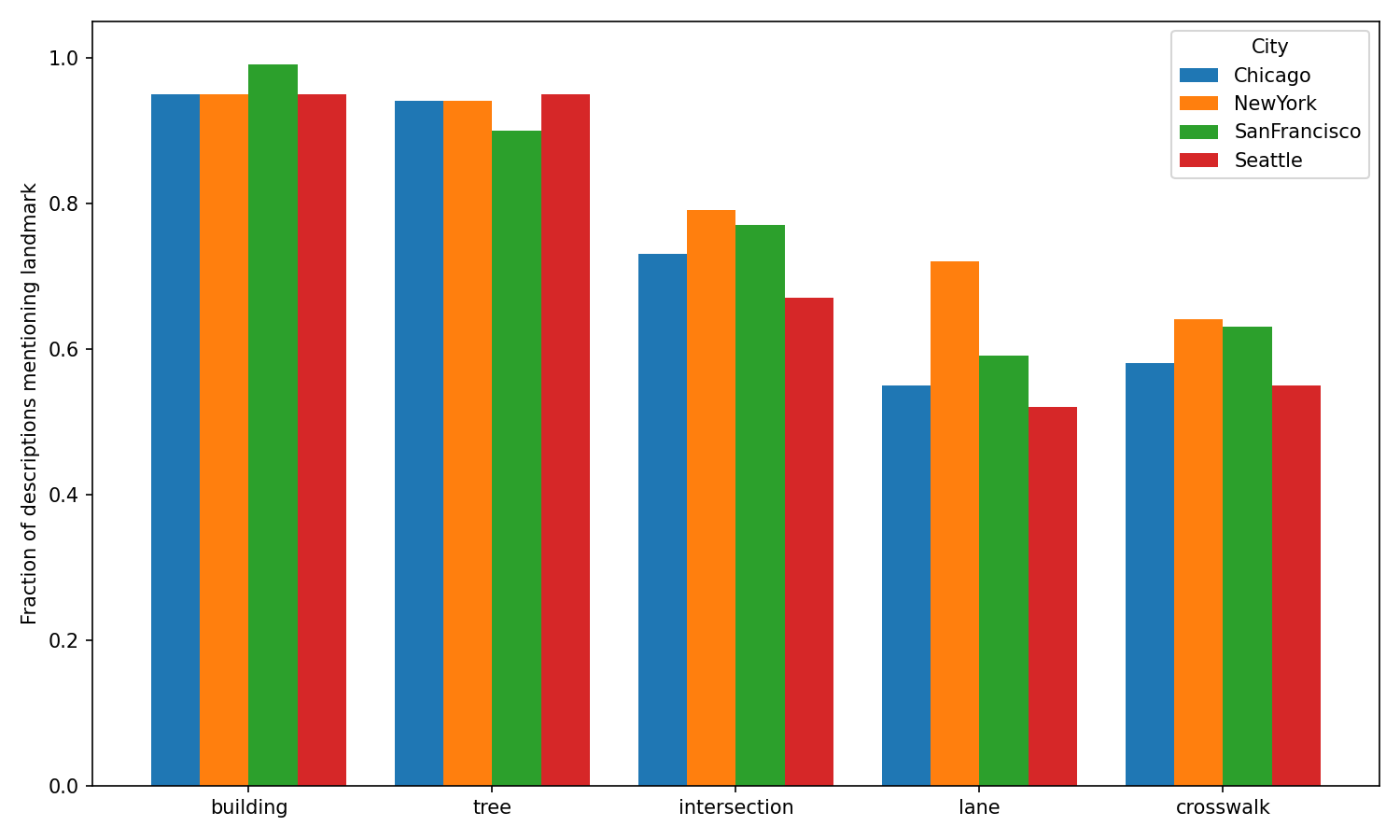}
        \caption{Ground panorama descriptions.}
    \end{subfigure}
    \hfill
    \begin{subfigure}{0.49\linewidth}
        \centering
        \includegraphics[width=\linewidth]{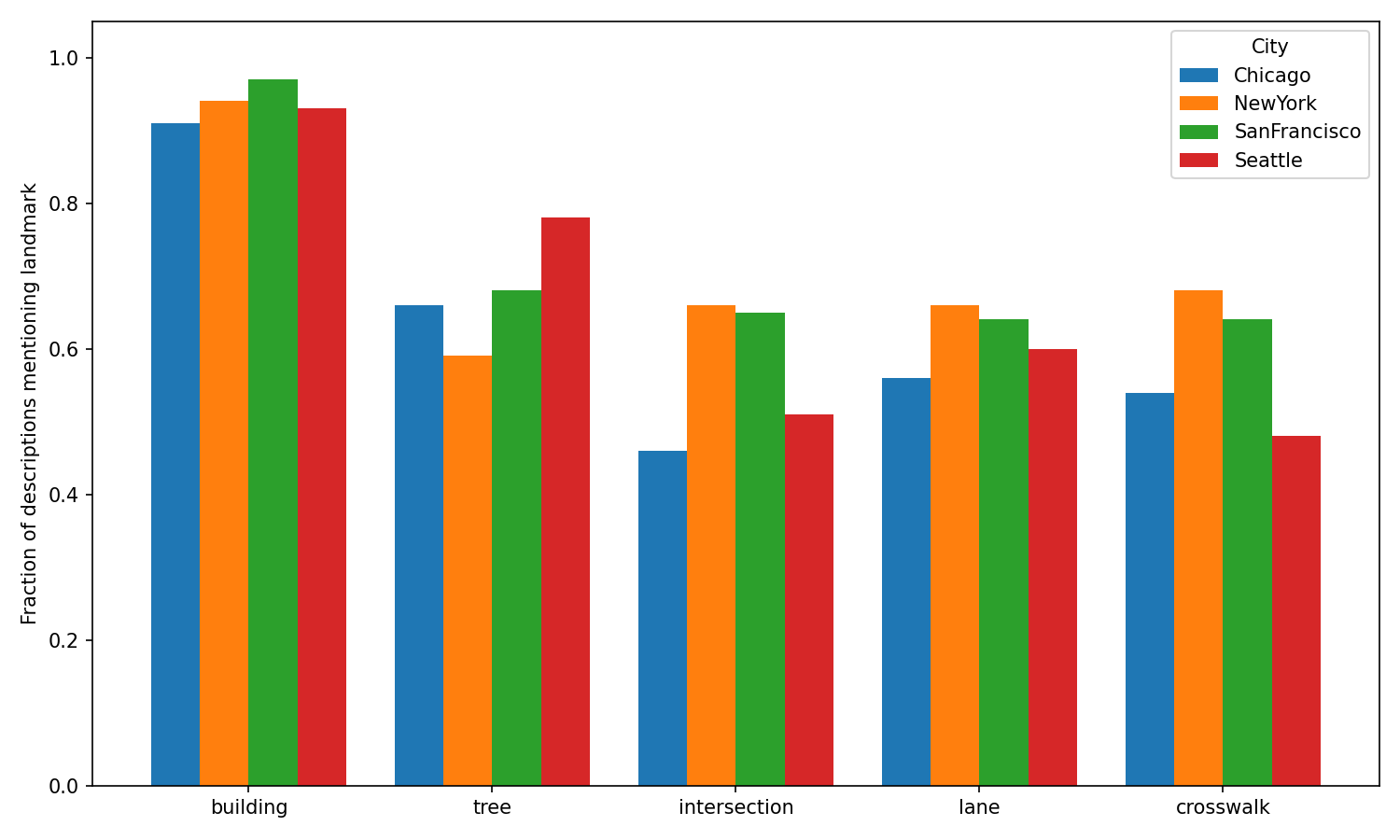}
        \caption{Satellite tile descriptions.}
    \end{subfigure}
    \caption{Frequency of semantic anchors across the four cities for both views, supporting Section~\ref{subsec:res_lexical}. Frequencies are consistent between views, and the dominant anchors are close to ceiling, so the terms shared most reliably across modalities are also the least useful for telling two locations apart.}
    \label{fig:bow}
\end{figure}

\begin{figure}[h!]
    \centering
    \includegraphics[width=1.0\linewidth]{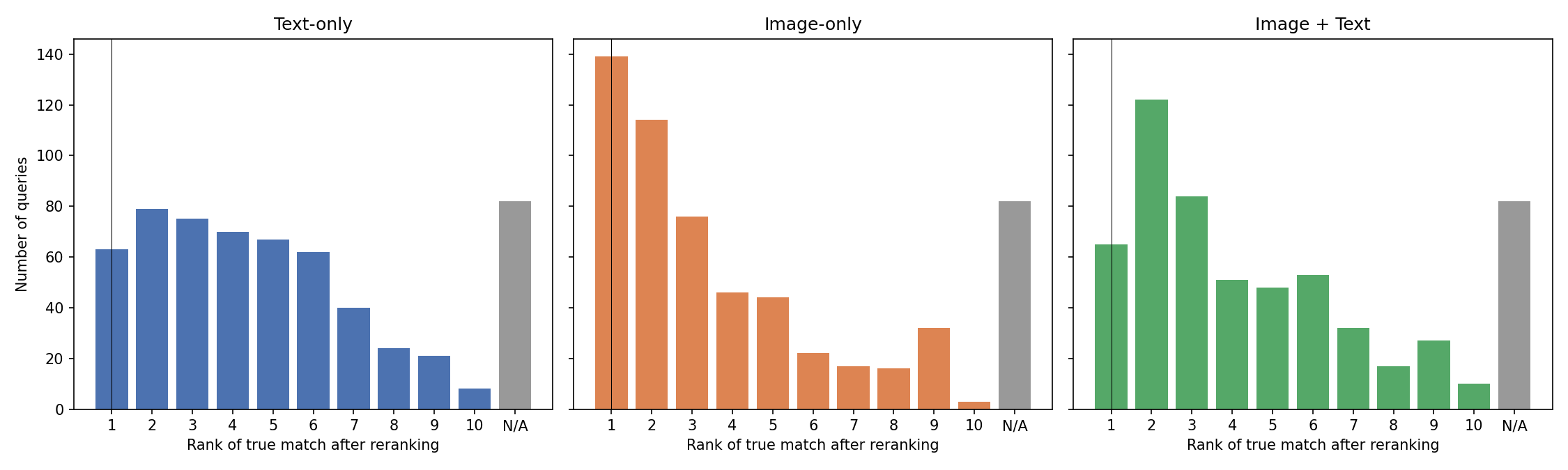}
    \caption{Distribution of the true tile's rank after listwise reranking on the hard subset, one panel per reranker input ("N/A" marks queries whose true tile was not in the retriever's top-10 at all). The image-only reranker is sharply peaked at rank 1, while the text-only and combined variants are close to flat, which confirms the chance-level Recall@1 in Figure~\ref{fig:rerank_bar} distributionally rather than through a single summary metric.}
    \label{fig:rerank_dist}
\end{figure}

This section collects the supporting figures referenced from the main text: dataset samples (Figure~\ref{fig:vigor_overview}), embedding similarity (Figure~\ref{fig:cosine}), lexical distributions (Figure~\ref{fig:bow}), and the post-reranking rank histograms (Figure~\ref{fig:rerank_dist}).

\section{Field Ablation and Grounding Audit}
\label{sec:appendix_ablation}

Table~\ref{tab:ablation} reports the field ablation discussed in Section~\ref{subsec:res_retrieval}. Each condition repeats the pointwise verification task on 120 queries drawn from the same hard negative pools, exposing only a subset of the schema to the judge: all fields; topology fields only (\texttt{road\_topology\_360}, \texttt{cardinal\_orientation}, \texttt{road\_geometry}); landmark fields only (\texttt{distinctive\_anchors}, \texttt{linear\_sequence}); or orientation fields only (\texttt{orientation\_cues}, \texttt{cardinal\_orientation}). Note that \texttt{cardinal\_orientation} appears in both the topology and the orientation conditions, since it is the satellite-side counterpart of each, so the two conditions are not strict complements. The all-fields condition on this subsample reproduces the full $n = 600$ run to within $0.2$ points, which serves as an internal consistency check on both runs. All conditions use the same query list and the same pools, so the comparison is paired even though we report unpaired intervals. Sample size should be kept in view when reading the table: at $n = 120$ the $95\%$ binomial interval on Recall@1 is roughly $\pm 7$ points, so the orientation-only condition is the only one that differs from the all-fields condition beyond sampling noise, and the ordering among the other three conditions should be read as suggestive.

\begin{table}[h!]
    \centering
    \small
    \caption{Field ablation for MLLM structured verification, $n = 120$ queries per condition on ten-candidate hard negative pools. Values are percentages except MRR. Road topology and geometry account for most of the signal, and orientation cues alone fall to or below the $10\%$ random baseline at Recall@1. With intervals of roughly $\pm 7$ points, only the orientation-only condition separates from all fields, so no entry is bolded.}
    \label{tab:ablation}
    \begin{tabular}{lcccc}
        \toprule
        \textbf{Fields exposed to the judge} & \textbf{R@1} & \textbf{R@3} & \textbf{R@5} & \textbf{MRR} \\
        \midrule
        All fields                      & 20.8 & 43.3 & 63.3 & 0.394 \\
        Road topology and geometry only & 17.5 & 48.3 & 70.0 & 0.385 \\
        Landmarks only                  & 15.0 & 36.7 & 64.2 & 0.350 \\
        Orientation cues only           & 5.8  & 23.3 & 50.0 & 0.257 \\
        \bottomrule
    \end{tabular}
\end{table}

For the grounding audit we sampled 80 ground panoramas and split their \texttt{distinctive\_anchors} field into 356 individual claims. Each claim was presented back to the MLLM together with its source image and labeled GROUNDED (clearly visible), PARTIAL (present but overstated or misdescribed), or HALLUCINATED (not visible at all), yielding $92.1\%$, $3.1\%$, and $4.8\%$ respectively. We target this field because it is the one the model is most free to invent, and the one whose fabrication would be most damaging in a navigation setting. Per-claim rationales are retained alongside the labels. Because the auditing model also generated the claims, self-preference is expected to push the GROUNDED share upward~\cite{zheng2023judging}: $92.1\%$ should therefore be read as an optimistic figure and $4.8\%$ as a floor on the true hallucination rate, and an audit by an independent model or by human annotators would be needed to estimate it. The audit also covers only \texttt{distinctive\_anchors}, not the topology fields that carry most of the verification signal.

\section{Qualitative Verifier Decisions}
\label{sec:appendix_qualitative}

The verification results in Section~\ref{subsec:res_retrieval} are reported as recall and MRR, which say how often the judge is right but not what its reasoning looks like. This section reproduces three complete decisions, drawn from the same $n = 600$ verification run: one correct match, one wrong candidate accepted, and one true match wrongly rejected. All text is verbatim, including the field labels the judge sees, and the images are the actual VIGOR panorama and tiles. Together they show the property we claim for this formulation, namely that every verdict comes with an inspectable argument, and they also show two distinct ways that argument goes wrong.

\subsection{A correct match}

\begin{figure}[h!]
    \centering
    \begin{subfigure}{0.63\linewidth}
        \centering
        \includegraphics[width=\linewidth]{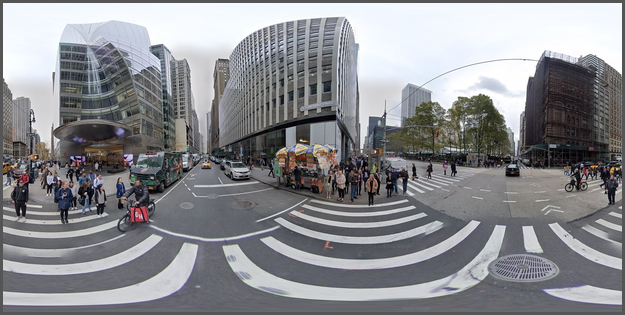}
        \caption{Ground panorama (query).}
    \end{subfigure}
    \hfill
    \begin{subfigure}{0.31\linewidth}
        \centering
        \includegraphics[width=\linewidth]{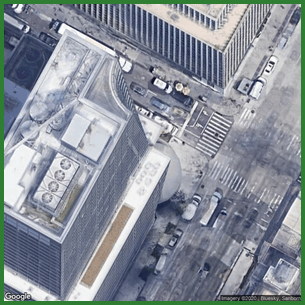}
        \caption{True match (satellite).}
    \end{subfigure}
    \caption{A correct verification on a dense Midtown Manhattan intersection, scored $9/10$.}
    \label{fig:qual_correct}
\end{figure}

Figure~\ref{fig:qual_correct} shows the query and the tile the verifier accepted.

\noindent\textbf{Ground fields shown to the verifier.}
{\small\begin{itemize}\setlength{\itemsep}{1pt}
    \item \textit{road topology 360:} The road network is continuous, forming a large four-way intersection. The road seen in the Center (Front) connects directly to the road visible at the Edges (Back), which wraps around the sphere to form a complete loop. The intersection is wide and complex, with multiple lanes and pedestrian crossings.
    \item \textit{spatial layout:} Left Side Sequence: Modern glass building $\rightarrow$ Food cart $\rightarrow$ Pedestrian crossing $\rightarrow$ Construction site. Right Side Sequence: High-rise office building $\rightarrow$ Pedestrian crossing $\rightarrow$ Street vendor $\rightarrow$ Park with trees.
    \item \textit{distinctive anchors:} The modern glass building with a curved entrance on the left, the food cart with a colorful umbrella in the center, the construction site with scaffolding on the right, and the large, curved building with a grid-like facade in the center.
    \item \textit{orientation cues:} The scene is a major urban intersection in a dense city. The architecture, street layout, and presence of a food cart suggest a typical New York City environment. The lack of strong shadows makes compass direction difficult to determine from lighting alone.
\end{itemize}}

\noindent\textbf{Satellite candidate (true match).}
{\small\begin{itemize}\setlength{\itemsep}{1pt}
    \item \textit{cardinal orientation:} North-South and East-West alignment of streets.
    \item \textit{road geometry:} 4-way intersection with crosswalks, multiple lanes of traffic, presence of a large parking lot adjacent to the road.
    \item \textit{linear sequence:} Commercial building with a curved corner, adjacent building with a flat roof, parking lot with multiple vehicles, road with crosswalks and traffic.
    \item \textit{distinctive anchors:} Intersection of 5th Avenue and W 40th Street, curved corner of 4 Times Square, circular structure at 1271 Avenue of the Americas, distinctive grid of streets in Midtown Manhattan.
\end{itemize}}

\noindent\fbox{\begin{minipage}{0.97\linewidth}\small
\textbf{Verifier output, score $9/10$, verdict MATCH.} "The ground-level description of a complex four-way intersection with a modern glass building, food cart, construction site, and high-rise office building aligns with the satellite view of the intersection of 5th Avenue and W 40th Street in Midtown Manhattan, including the curved corner of 4 Times Square and the grid layout."
\end{minipage}}

\vspace{0.6em}
The verdict is correct and the argument is checkable line by line, which is what the interpretability claim in Section~\ref{subsec:res_retrieval} amounts to in practice. The example also exposes a confound worth naming. The satellite description does not merely describe geometry, it names 5th Avenue, 4 Times Square, and 1271 Avenue of the Americas, so the model has recognized a famous location from world knowledge, and the verifier's rationale then rests partly on that recognition rather than on cross-view comparison alone. In landmark-dense city centers, some share of our verification accuracy may come from memorized geography. We expect this to be a small effect overall, since the great majority of our tiles are ordinary residential and commercial blocks whose descriptions contain no proper nouns at all, but it is a reason to prefer generic scenes when reading individual successes.

\subsection{A wrong candidate accepted}

\begin{figure}[h!]
    \centering
    \begin{subfigure}{0.55\linewidth}
        \centering
        \includegraphics[width=\linewidth]{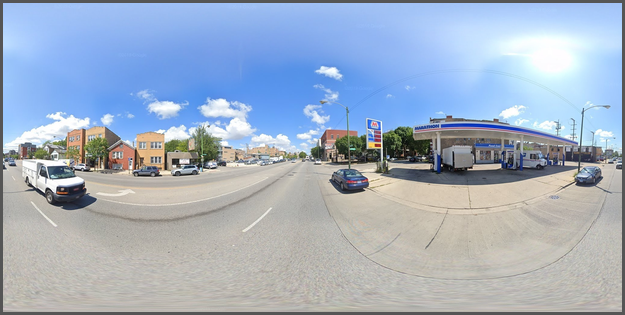}
        \caption{Ground panorama (query).}
    \end{subfigure}
    \hfill
    \begin{subfigure}{0.26\linewidth}
        \centering
        \includegraphics[width=\linewidth]{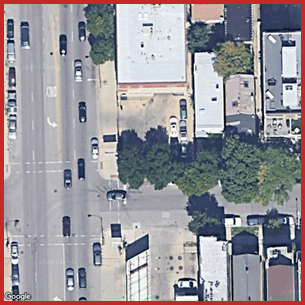}
        \caption{Accepted, wrong.}
    \end{subfigure}
    \hfill
    \begin{subfigure}{0.16\linewidth}
        \centering
        \includegraphics[width=\linewidth]{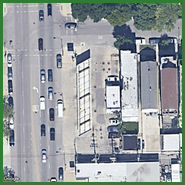}
        \caption{True match.}
    \end{subfigure}
    \caption{A false accept, scored $6/10$. The accepted tile and the true tile are both ordinary four-lane commercial blocks in the same city.}
    \label{fig:qual_falsematch}
\end{figure}

Figure~\ref{fig:qual_falsematch} shows the query, the tile the verifier accepted, and the true tile it passed over.

\noindent\textbf{Ground fields shown to the verifier.}
{\small\begin{itemize}\setlength{\itemsep}{1pt}
    \item \textit{road topology 360:} A continuous, multi-lane road runs straight ahead and curves around to the left and right, forming a T-intersection. The road continues forward and to the left (left edge), and the road behind the camera (right edge) connects to the same intersection.
    \item \textit{spatial layout:} Left Side Sequence: Residential buildings with brick facades, a small commercial building, parked cars, and a white van. Right Side Sequence: Marathon gas station with fuel pumps, a white delivery truck, and a white van parked at the pumps.
    \item \textit{distinctive anchors:} Marathon gas station with a blue and white canopy, white delivery truck at the gas pumps, red brick residential buildings on the left, and a white van on the left side of the road.
    \item \textit{orientation cues:} The sun is high in the sky, casting short shadows towards the left side of the image, indicating the sun is in the east, suggesting it is morning. The orientation of the road and the position of the gas station suggest a northeast heading.
\end{itemize}}

\noindent\textbf{Satellite candidate (accepted, wrong).}
{\small\begin{itemize}\setlength{\itemsep}{1pt}
    \item \textit{cardinal orientation:} Road runs approximately North-South.
    \item \textit{road geometry:} 4-lane road with clear lane markings, turn lanes at intersection, visible crosswalks, sidewalks along both sides, multiple parked cars along the street.
    \item \textit{linear sequence:} Road with multiple lanes, adjacent parking lot, large building with flat roof, row of trees, residential houses with varying roof types, driveways, parked cars.
    \item \textit{distinctive anchors:} 4-way intersection with visible turn lanes, large flat-roofed commercial building, rectangular parking lot with cars, linear row of trees separating street from property, unique white building with a flat roof, distinct pattern of parked cars along the street.
\end{itemize}}

\noindent\fbox{\begin{minipage}{0.97\linewidth}\small
\textbf{Verifier output, score $6/10$, verdict MATCH.} "Both descriptions feature a multi-lane road with a T-intersection, a Marathon gas station on the right, residential buildings on the left, and a white van; the road orientation and morning sunlight suggest consistent alignment, though satellite details like the flat-roofed building and tree row are not confirmed in the ground-level view."
\end{minipage}}

\vspace{0.6em}
Two failures are visible here, and they are of different kinds. The first is the discriminability problem this paper is about: every property the two descriptions genuinely share, a multi-lane road, a T-intersection, parked cars, mixed residential and commercial frontage, is shared by most commercial blocks in the same city, so the evidence is real but not distinguishing. The second is a fault in the rationale itself. The judge asserts that both descriptions mention a Marathon gas station, and the satellite description mentions no gas station at all; the agreement is invented. Note also that the judge does register the disconfirming evidence, since it observes that the flat-roofed building and the tree row are not confirmed from the ground, and returns MATCH regardless, which suggests the pointwise scoring rubric weights agreement more heavily than contradiction. This is the concrete form of the hallucination risk discussed in Section~\ref{subsec:limitations}, and it is also the argument for the human-in-the-loop deployment we recommend, since the error is legible in the rationale even though it is invisible in the score.

\subsection{A true match wrongly rejected}

\begin{figure}[h!]
    \centering
    \begin{subfigure}{0.63\linewidth}
        \centering
        \includegraphics[width=\linewidth]{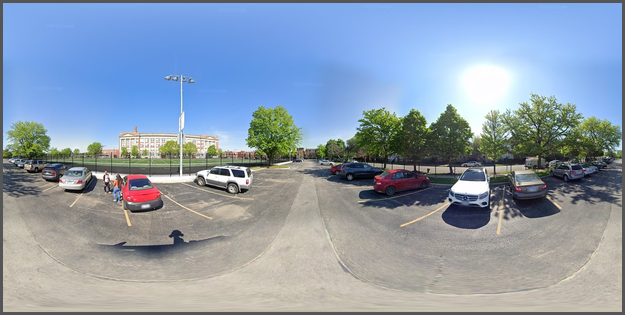}
        \caption{Ground panorama (query).}
    \end{subfigure}
    \hfill
    \begin{subfigure}{0.31\linewidth}
        \centering
        \includegraphics[width=\linewidth]{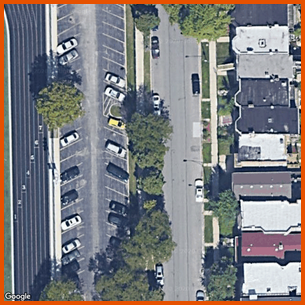}
        \caption{True match, rejected.}
    \end{subfigure}
    \caption{A false reject, scored $2/10$. The panorama sits in a parking lot at the edge of its tile, so the two views cover overlapping but largely different ground.}
    \label{fig:qual_falsereject}
\end{figure}

Figure~\ref{fig:qual_falsereject} shows the query and the true tile the verifier rejected.

\noindent\textbf{Ground fields shown to the verifier.}
{\small\begin{itemize}\setlength{\itemsep}{1pt}
    \item \textit{road topology 360:} Continuous asphalt parking lot with a single, wide, curved road running through the center, connecting the front and back of the scene. The road is a dead-end at the far end, where it meets a fence and a building.
    \item \textit{spatial layout:} Left Side Sequence: Parking lot $\rightarrow$ Black fence $\rightarrow$ Green football field $\rightarrow$ Brick school building. Right Side Sequence: Parking lot $\rightarrow$ Trees $\rightarrow$ Black fence $\rightarrow$ Brick school building.
    \item \textit{distinctive anchors:} Red car parked in the foreground, white SUV parked in the foreground, brick school building, green football field, tall light pole.
    \item \textit{orientation cues:} The sun is high in the sky, indicating midday. The shadows of the cars and the light pole are short and fall towards the left side of the image, suggesting the camera is facing approximately south.
\end{itemize}}

\noindent\textbf{Satellite candidate (true match, rejected).}
{\small\begin{itemize}\setlength{\itemsep}{1pt}
    \item \textit{cardinal orientation:} North-South.
    \item \textit{road geometry:} Two-lane road with a central median strip, no visible lane markings or turn arrows, straight alignment.
    \item \textit{linear sequence:} Continuous asphalt track, long row of parked cars, narrow grassy strip, paved road with central median, continuous row of houses with front lawns and driveways.
    \item \textit{distinctive anchors:} Running track and adjacent parking lot, parking lot with distinct yellow car, unique row of houses with varied roof colors, long straight road with central median.
\end{itemize}}

\noindent\fbox{\begin{minipage}{0.97\linewidth}\small
\textbf{Verifier output, score $2/10$, verdict NO\_MATCH.} "The ground-level description shows a cul-de-sac with a dead-end and a school building, while the satellite view shows a straight road with a central median and houses, indicating incompatible road topology and layout."
\end{minipage}}

\vspace{0.6em}
This failure is the mirror image of the previous one, and it is not a reasoning error. Given the two descriptions, the verdict follows: a dead-end parking lot beside a school and a straight road with a central median and a row of houses are genuinely incompatible layouts. The premise is what fails. Because VIGOR panoramas are not centered on their tiles (Figure~\ref{fig:vigor_overview}), the camera here stands in a parking lot near the edge of its tile, and the two descriptions faithfully report overlapping but largely different ground: the panorama is dominated by the lot, the running track, and the school, while the tile is dominated by the road and housing to one side. Both descriptions do share the track and the adjacent parking lot, but that agreement is buried among the disagreements produced by the offset. Description-based verification has no mechanism for partial overlap, since it compares two accounts as though they should describe the same content, which points to a concrete fix outside this paper: conditioning the judge on the expected offset, or asking for agreement over the overlapping sector rather than over the scene as a whole.

\section{Implementation Details}
\label{sec:appendix_impl}

The 9{,}826 schema-valid pairs recovered by the repair module are distributed evenly across the four cities: Chicago 2{,}469, New York 2{,}464, San Francisco 2{,}450, and Seattle 2{,}443. All description generation and all LLM verification and reranking calls use Qwen 3 VL 30B A3B Instruct with greedy decoding, served on NVIDIA H200 GPUs. Descriptions were generated once per image and cached; every downstream experiment reads from that cache, so the text-based methods are compared on byte-identical inputs.

The verification experiment issues one LLM call per candidate, giving 6{,}000 calls for the $n = 600$ pointwise run and 1{,}200 per ablation condition. The reranking experiment issues one call per query, presenting all ten candidates in a single prompt; we verified that the model's chat template accepts multiple labeled images in one turn, with each image receiving its own vision block. Reranking runs at roughly 30 to 55 seconds per query depending on the input variant and the node, and the three variants were run as separate jobs over identical query lists with the same random seed.

For the visual front-end we use a released AuxGeo-family ConvNeXt-base checkpoint trained on the VIGOR same-area split, without any further training or adaptation on our part, and we compute embeddings for the panoramas and for every tile in our reduced pool before taking the top-10 by cosine similarity. Before the main run we sanity-checked the checkpoint on a handful of hand-picked pairs, obtaining true-pair similarities of $0.54$ to $0.62$ against wrong-pair similarities of $0.13$ to $0.20$, and then validated it at scale against our pool as reported in Section~\ref{subsec:res_rerank}.

\end{document}